%% file: arxiv_main.tex
\documentclass{article}

\usepackage{microtype}
\usepackage{graphicx}
\usepackage{subcaption}
\usepackage{booktabs}
\usepackage{transparent}

\usepackage{hyperref}
\usepackage{soul}
\usepackage{graphicx}
\usepackage{tcolorbox}
\usepackage{xurl}

\usepackage[accepted]{paper_sty}

\usepackage{tabularx}
\tcbuselibrary{breakable}

\usepackage{amsmath}
\usepackage{amssymb}
\usepackage{mathtools}
\usepackage{amsthm}
\usepackage{xspace}
\usepackage{makecell}

\usepackage[capitalize,noabbrev]{cleveref}
\usepackage{scalefnt}

\theoremstyle{plain}

\theoremstyle{definition}

\theoremstyle{remark}

\input{algo}

\usepackage[textsize=tiny]{todonotes}

\icmltitlerunning{\method}

\begin{document}

\twocolumn[
\icmltitle{\method: Automated Training Framework of Universal Process Reward Models via Ensemble Prompting and Reverse Verification}

\icmlsetsymbol{equal}{*}
\icmlsetsymbol{corr}{†}

\begin{icmlauthorlist}
\icmlauthor{Xiaoyu Tan}{equal,INF}
\icmlauthor{Tianchu Yao}{equal,INF}
\icmlauthor{Chao Qu}{equal,INF}
\icmlauthor{Bin Li}{sues}
\icmlauthor{Minghao Yang}{Fudan}
\icmlauthor{Dakuan Lu}{INF}
\icmlauthor{Haozhe Wang}{INF}
\icmlauthor{Xihe Qiu}{corr,sues}
\icmlauthor{Wei Chu}{INF}
\icmlauthor{Yinghui Xu}{Fudan}
\icmlauthor{Yuan Qi}{Fudan}
\end{icmlauthorlist}

\icmlaffiliation{INF}{INFLY TECH (Shanghai) Co., Ltd., Shanghai, China}
\icmlaffiliation{Fudan}{Fudan University, Shanghai, China}
\icmlaffiliation{sues}{Shanghai University of Engineering Science, Shanghai, China}

\vskip 0.3in
]

\printAffiliationsAndNotice{\icmlEqualContribution \icmlEqualCorr} 

\begin{abstract}
The reasoning capabilities of advanced large language models (LLMs) like o1 have revolutionized artificial intelligence applications. Nevertheless, evaluating and optimizing complex reasoning processes remain significant challenges due to diverse policy distributions and the inherent limitations of human effort and accuracy. In this paper, we present \methodr\footnote{The project has been open-sourced at \href{https://auroraprm.github.io/}{auroraprm.github.io}}, a novel automated framework for training universal process reward models (PRMs) using ensemble prompting and reverse verification. The framework employs a two-phase approach: First, it uses diverse prompting strategies and ensemble methods to perform automated annotation and evaluation of processes, ensuring robust assessments for reward learning. Second, it leverages practical reference answers for reverse verification, enhancing the model's ability to validate outputs and improving training accuracy. To assess the framework's performance, we extend beyond the existing ProcessBench benchmark by introducing UniversalBench, which evaluates reward predictions across full trajectories under diverse policy distribtion with long Chain-of-Thought (CoT) outputs. Experimental results demonstrate that \methodr enhances process evaluation accuracy, improves PRMs' accuracy for diverse policy distributions and long-CoT responses. The project will be open-sourced at \href{https://auroraprm.github.io/}{auroraprm.github.io}. The \UNI is available at \href{https://huggingface.co/infly/Universal-PRM-7B}{huggingface.co/infly/Universal-PRM-7B}.

\end{abstract}

\section{Introduction}

The rapid development of large language models (LLMs) has highlighted their potential as foundational components of artificial general intelligence (AGI), driven by advancements in scaling laws for model size and inference \cite{openai2023gpt, park2023generative, kaddour2023challenges, zhu2024deepseek, zheng2023judging}. Inference scaling laws \cite{muennighoff2023scaling, wei2022emergent} demonstrate that allocating additional computational resources during the inference phase of LLMs, rather than solely increasing model size, can substantially improve accuracy and problem-solving capabilities. Recent advances \cite{besta2024graph, yao2024tree, zhao2024large, shinn2024reflexion, koa2024learning} exemplify this by incorporating mechanisms such as Chain-of-Thought (CoT) \cite{wei2022cot, wang2022self} reasoning for structured intermediate steps, Monte Carlo Tree Search (MCTS) \cite{kocsis2006bandit, coulom2006efficient, swiechowski2023monte} for strategic exploration of solution paths, and reflection processes \cite{ji2023towards, shinn2024reflexion} for iterative self-improvement. These mechanisms can be learned by multiple post-training methods, such as reinforcement learning (RL) from various source feedback and supervised fine-tuning \cite{wang2024math, teamqwen2024qwq, openai2024o1}. These methods enable stepwise reasoning and systematic search during LLM inference, underscoring the importance of optimizing inference methods alongside model scaling to enhance generation capabilities and achieve advanced reasoning performance in real-world complex scenarios.

Generating and learning long reasoning sequences pose significant challenges, as they require robust methods to evaluate and ensure the suitability or correctness of each reasoning step or generated segment. Recent research suggests that process reward models (PRMs) can be trained and employed to verify individual reasoning steps generated by LLMs \cite{lightman2023let, luo2024improve}. Unlike outcome reward models (ORMs) \cite{cobbe2021orm1, yu2023orm2}, which provide a singular evaluation at the end of a reasoning sequence, PRMs offer dense reward signals throughout the sequence by assessing each reasoning step. This granularity allows PRMs to determine whether a given step is steering the reasoning process toward the correct outcome. The ability of PRMs to provide step-by-step evaluations can be directly applied in guided search, where they work in conjunction with various search algorithms to generate reasoning sequences more effectively \cite{wang2024math, lightman2023let}. Additionally, PRMs can serve as a feedback mechanism for RL algorithms \cite{zhang2025lessons}, supplying intermediate rewards that significantly enhance the convergence efficiency, robustness, and performance of the learned policy. By ensuring the correctness of both the process and the final outcome, PRMs are a pivotal advancement in enhancing the reliability and effectiveness of LLMs' reasoning capabilities.

However, achieving high accuracy in PRM training presents significant challenges, primarily due to the difficulty of constructing high-quality datasets \cite{uesato2022solving}. Ensuring data quality often involves human annotators, who label the data to provide ground truth. While instruction-level annotation focuses on assessing the overall quality of a response \cite{lightman2023let}, process-level annotation requires annotators to evaluate each reasoning step and determine whether it leads to the correct outcome. This step-by-step labeling demands not only expert knowledge but also substantially more human resources, making the process both resource-intensive and time-consuming. To address these challenges, researchers have explored automated annotation methods, such as rolling out and calculating stepwise accuracies \cite{zheng2024processbench, wang2024math} or prompting instruct LLMs for self-verification \cite{cao2024towards, zhou2024your, fu2022complexity}. Although promising, these approaches are typically operated and optimized based on the target policy and focus on finding the first error location, which can restrict the versatility of the PRM in evaluating a wide range of policies, experience performance degradation when applied to out-of-distribution (OOD) policies, and reduce usability in optimizing subsequent RL algorithms that require complete process rewards along the trajectory. These limitations significantly limit their universally generalization. From a reinforcement learning (RL) perspective, these methods are training a partially policy reward function under specific sampling policies with partial state distribution which is not designed to be used universally.  In addition to these limitations, we also identify an underutilization of reference information in current PRMs. Leveraging reference information could significantly enhance PRM performance, thereby improving subsequent policy learning. We found that addressing these gaps is essential for advancing the performance of PRMs in universal policy evaluation scenarios.

In this paper, we introduce \methodr, a novel \textbf{A}utomated training framework for \textbf{U}niversal P\textbf{R}Ms that leverages ensemble pr\textbf{O}mpting and \textbf{R}everse verific\textbf{A}tion. The proposed framework includes several key stages: we first generate diverse responses using various LLMs as base policy to ensure broad coverage of output policy distributions. Next, various prompting strategies guide instruction-based LLMs to act as discriminators, evaluating each reasoning step to determine whether the candidate steps navigate to the correct answer. By combining the outputs of multiple prompting strategies through an ensemble approach, prediction accuracy and robustness are significantly enhanced. Finally, the ensemble-labeled data, optionally combined with reference answers, is used to train PRMs, which can enable reverse verification and further improve accuracy by cross-validating predictions against available reference answers. This learning process enable the training of a universal PRM with high generalization capabilities in predicting on various policy distributions and entire reasoning trajectories. To thoroughly assess the effectiveness of our proposed method, we conduct evaluations on ProcessBench \cite{zheng2024processbench} and introduce a new benchmark, UniversalBench. UniversalBench is designed to include long CoT outputs and enable a comprehensive evaluation of the entire reasoning process for each question. Our experimental results demonstrate that \methodr achieves strong performance and generalization. Ablation studies and detailed analyses reveal that ensemble prompting and reverse verification with reference answers substantially enhance PRM accuracy and generalization. Key contributions of our work include:
\begin{itemize}
    \item We propose an ensemble prompting method that aggregates results from multiple LLMs, demonstrating strong performance while significantly reducing the need for human labeling.
    \item We introduce reverse verification using optional reference answers during training and inference, which significantly improves PRM accuracy. 
    \item We design UniversalBench to assess the capability of PRMs in evaluating long CoT outputs and providing predictions throughout the entire reasoning sequence. This design aligns more closely with the practical scenarios of PRMs in RL settings.
    \item Our framework achieves superior performance on two benchmarks. We have provided open access to our trained PRM \UNI and will release UniversalBench for community use\footnotemark[\value{footnote}].
\end{itemize}

\section{Related Works}
Reasoning capability is one of the most important perspectives where the LLMs are considered as the key step to the artificial general intelligence. Multiple reasoning policies are proposed to improve the reasoning abilities of LLMs, aiming to activate their reasoning capabilities and make them more interpretable and efficient in solving reasoning problems that require multi-step inference and complex reasoning. One of the earliest methods to improve reasoning capabilities was the Chain of Thought (CoT) \cite{wei2022cot} prompting, encouraging LLMs to generate intermediate reasoning steps explicitly. However, it still relies on a relatively simple, linear flow of thought, which can become limiting for tasks involving more complex reasoning. To address this limitation, the Tree of Thought (ToT) \cite{yao2024tree} extends CoT by organizing the reasoning process into a tree-like structure, enabling LLMs to consider multiple reasoning paths and self-evaluate the next step. Recent studies \cite{yu2023metamath, luo2023wizardmath, yue2023mammoth, gou2023tora} developed high-quality math reasoning step datasets using methods such as CoT and ToT to fine-tune LLMs and enhance their reasoning capabilities. These methods enhance LLMs' mathematical reasoning potential by improving the quality of reasoning steps. The O1 Model \cite{openai2024o1, openai2024o1_mini} further boosts reasoning capability through test-time scaling laws \cite{snell2024scaling}, which provide the LLMs with more reasoning tokens during inference. Test-time scaling laws increase in reasoning tokens enhances the model’s overall reasoning capacity, enabling it to tackle more complex tasks.

In addition to activating the reasoning capability of LLMs through prompts or high-quality reasoning datasets, recent studies \cite{zheng2024processbench, zhang2025lessons, mcaleese2024llm} found that constraining LLMs to generate answers from multiple decoding candidates by reward models can also enhance LLMs reasoning capacity. There are two types of reward models: the Process Reward Model (PRM) and the Outcome Reward Model (ORM). ORM evaluates the whole reasoning process based on the final answer, ignoring intermediate steps. In contrast, PRM assesses the quality of each reasoning step individually. However, LLMs frequently make computational or logical errors in mathematical reasoning tasks. Even when producing correct answers, LLMs may generate plausible but incorrect reasoning steps, compromising the reliability of the models \cite{wang2024math}. Employing reward models to evaluate each step of the reasoning process, enhances the reasoning capability and generative reliability of LLMs \cite{lightman2023let}. PRM has been proven superior to ORM \cite{wu2023fine}, as it not only evaluates the reasoning process but also provides reward signals for each individual reasoning step \cite{uesato2022solving, pan2023let}. This helps generate higher-quality data, thereby strengthening the reasoning capability of LLMs. However, training PRM demands high-quality data, which presents significant challenges in terms of data annotation \cite{luo2024improve}. Recent work suggests that combining the LLM-as-a-judge \cite{zheng2023judging} framework with Monte Carlo estimation provide more accurate and consistent annotations for training PRM, thus enhancing the quality of the data for model training \cite{zhang2025lessons}.

\vspace{-0.08cm}
\section{Methods}
In this section, we first introduce the preliminary concepts of zero-shot prompting, in-context learning and ensemble learning. Then we introduce our proposed \methodr to train universal PRMs using ensemble prompting and reverse verification in an automated manner.

\begin{figure*}[ht]
\begin{center}
\centerline{\includegraphics[width=0.85\textwidth]{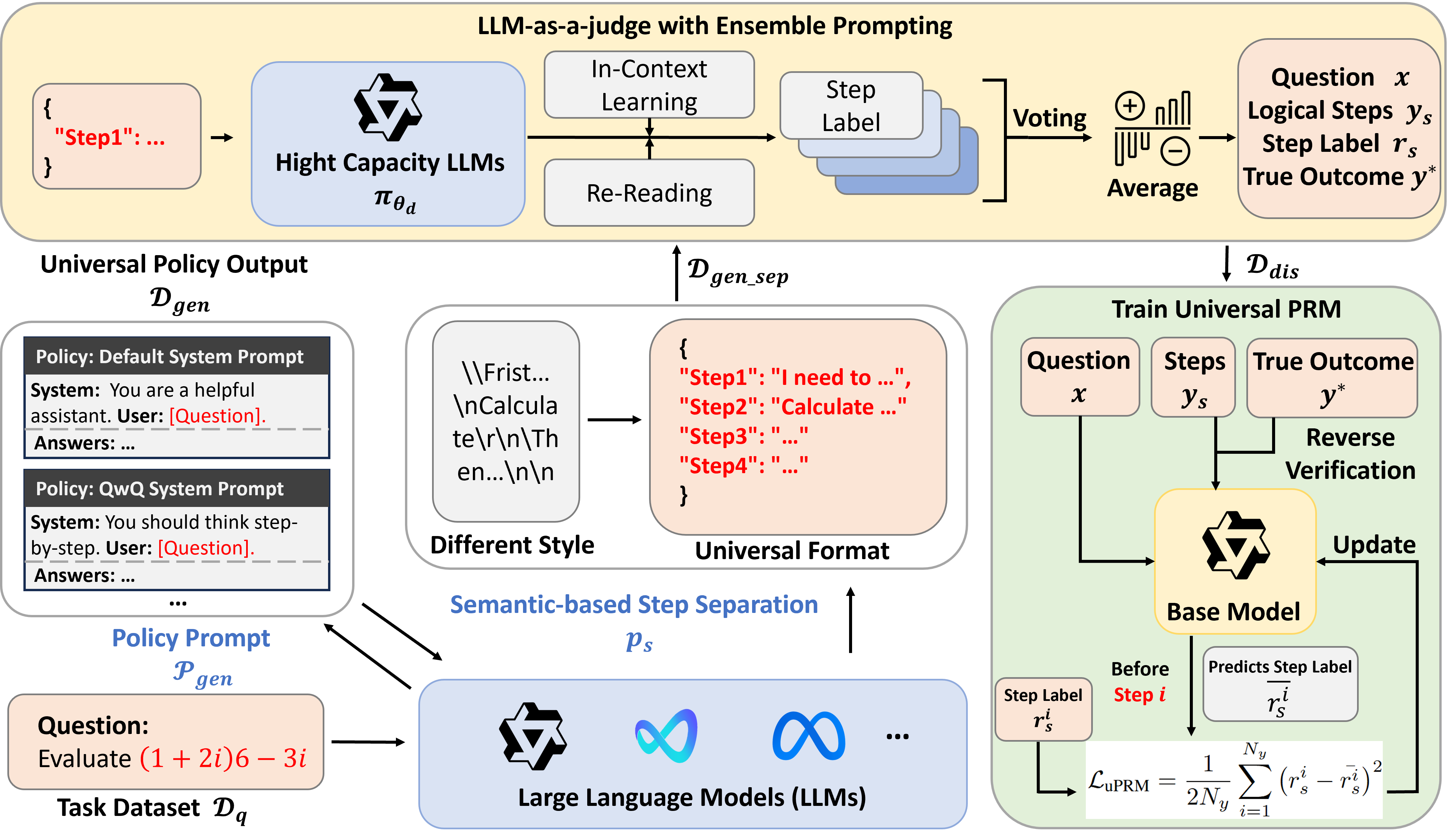}}
\caption{The Overall Workflow of \methodr.}
\label{framework}
\end{center}
\end{figure*}

\subsection{Preliminary}
\subsubsection{Zero-shot prompting and in-context learning}
Zero-shot prompting enables LLMs to perform tasks by utilizing task-specific instructions without task-specific fine-tuning. Formally, given an instruction fine-tuned LLM represented as \( f_\theta \) with parameters \( \theta \), it generates outputs based on the conditional probability \( y \sim f_\theta(\cdot|x, p) \), where \( p \) defines the task prompt and \( x \) is the input. This approach leverages the pre-trained knowledge and alignment obtained during post-training to generalize across a wide variety of tasks effectively. In-context learning, on the other hand, adapts LLMs to tasks by conditioning on a sequence of examples within the context window, rather than requiring model fine-tuning. Given \( \mathcal{D} = \{(x_i, y_i)\}_{i=1}^N \), the model predicts \( \hat{y}_{N+1} \sim f_\theta(\cdot|\mathcal{D}, x_{N+1}) \), extracting task-relevant patterns directly from the context. This paradigm facilitates few-shot learning and task adaptation using the representations learned during pretraining. The prompt engineering that utilize the combination of zero-shot prompting and in-context learning further enhances LLM capabilities, leveraging both natural language instructions and contextual adaptation to achieve robust task performance. Recent advancements \cite{jin2024impact, dong2024survey} highlight their synergy, enabling scalable, flexible task-solving through effective prompt design and context engineering.

\subsubsection{ORMs vs. PRMs}
The ORMs and PRMs represent two complementary approaches to reward modeling for evaluating the problem-solving processes of LLMs, differing in granularity and dataset requirements. ORM $R_{ORM}$ assigns a single real-valued score \( r_y = R_{ORM}(x,y) \) to a solution \( y \), indicating the likelihood that \( y\) is correct under question \(x\), using a cross-entropy loss to optimize the model defined as:  
\[
\mathcal{L}_{ORM} = -\mathbb{E}_i\left[ r_i \log \bar{r_i} + (1 - r_i) \log (1 - \bar{r_i}) \right],
\] where \( r \in \{0, 1\} \) is the ground truth label (\( r = 1 \) for correct solutions), and \( \bar{r} \in [0, 1] \) is the sigmoid-transformed prediction of ORMs. ORMs' training dataset, \( \mathcal{D}_{ORM} = \{(x_j, y_j, r_{j})\}_{j=1}^N \), includes \( N \) reasoning problems \( x_j \), candidate solutions \( y_j \) generated by LLMs, and binary correctness labels \( r_{j} \). While efficient and scalable, ORM's coarse-grained evaluations may misclassify solutions in complex reasoning tasks and provide sparse reward signals for subsequent policy leanring. PRMs $R_{PRM}$ enhances ORMs by providing fine-grained feedback through step-wise evaluation of solutions. Its loss function extends ORM to reasoning steps:  
\begin{equation}
\label{equ:prm_normal}
\mathcal{L}_{PRM} = -\mathbb{E}_{i,j} \left[ r_{s}^{j,i} \log \bar{r_s^{j,i}} + (1 - r_{s}^{j,i}) \log (1 - \bar{r_s^{j,i}}) \right],
\end{equation} 
where \( K \) denotes the number of reasoning steps in \( y \), \( r_{s}^i \in \{0, 1\} \) is the ground truth label for step \( i \), and \( \bar{r_s^{i}} = R_{PRM}(x,y_i) \) is the sigmoid-transformed prediction for step \( i \). PRM's dataset, \( \mathcal{D}_{PRM} = \{(x_j, \{y_{j,i}, r_{s}^{j,i}\}_{i=1}^{K_j})\}_{j=1}^N \), involves \( N \) problems \( x_j \), solutions decomposed into \( K_j \) steps \( \{y_{j,i}\}_{i=1}^{K_j} \), and step-wise correctness labels \( r_{s}^{j,i} \). While PRM offers more detailed and accurate feedback, it is less scalable due to the high cost of manual annotations required for step-wise correctness. Recent research has introduced various automated methods to enhance labeling efficiency. However, these approaches heavily depend on the sampling policy, making it challenging to generalize across different output policies and improve their performance \cite{lightman2023let, wang2024math, xiong2024building}. From a reinforcement learning perspective, these methods effectively train a policy reward function \( R_{PRM}^\pi \) under a specific sampling policy \( \pi \). This results in a reward function that lacks universality and is not suitable for facilitating improvements across all policies.

\subsection{\method}
\subsubsection{Universal Policy Output Generation}
\label{sec:universal_policy_output_generation}
To construct a universal PRM, \methodr begins by generating candidate policies from a diverse policy set \(\Pi = \{\pi_i\}_{i=1}^N\), comprising \(N\) LLM-based policies with varying output distributions. To further enhance the diversity of these distributions from different policy bases, we design a prompt set \(\mathcal{P}_{\text{gen}} = \{p_j\}_{j=1}^L\), consisting of \(L\) distinct instructions that specify various methods for generating target answers (Appendix \ref{app:p_gen}). Given a task dataset \(\mathcal{D}_{q} = \{(x_k, y^\star_k)\}_{k=1}^M\), containing \(M\) queries \(x_k\) paired with their correct answers \(y^\star_k\), it is important to note that the correct answers do not need to include detailed solving steps; only the final results are required.  The sample generation process systematically applies each policy \(\pi_i \in \Pi\) with every prompt \(p_j \in \mathcal{P}\) to all queries \(x_k \in \mathcal{D}_q\), resulting in a comprehensive set of outputs which can cover a wide range of output policy distribution. The resulting dataset of generated samples can be expressed as:
\begin{equation}
\mathcal{D}_{\text{gen}} = \bigcup_{i=1}^N \bigcup_{j=1}^L \bigcup_{k=1}^M \{(x_k, y_{i,j,k}\sim\pi_i(\cdot|p_j(x_k)),y^\star_k)\},
\end{equation} 
where \(y_{i,j,k}\sim\pi_i(\cdot|p_j(x_k))\) represents the response generated by policy \(\pi_i\) when applied to query \(x_k\) under prompt \(p_j\). This process ensures extensive coverage of the output distribution space, enabling robust training and evaluation of the universal PRM.

\subsubsection{Ensemble Prompting}
\label{sec:ensemble_prompting}

To enable autonomous labeling of process rewards, we employ the LLM-as-a-judge \cite{zheng2023judging} technique that utilizes LLMs as discriminators to generate a reward list for candidate answers based on the given question and reference answer. Recognizing the potential for inductive bias and variance in LLM outputs, we incorporate a majority voting mechanism to mitigate generation variance. Additionally, we use multiple prompting strategies to assess candidate answers from diverse inductive perspectives and apply ensemble techniques to further reduce bias and enhance reliability.

Before proceeding with the discrimination process, we first perform step-wise decomposition of all candidate responses in the dataset \(\mathcal{D}_{\text{gen}}\) by prompting the LLM \(\pi_{\theta_s}\) with parameters \(\theta_s\) using the prompt \(p_s\) (Appendix \ref{app:p_s}):
\[
    \mathcal{D}_{\text{gen\_sep}} = \bigcup_{ \mathcal{D}_{\text{gen}}} \{(x, y, y^\star, y_s \sim \pi_{\theta_s}(\cdot \mid p_s(x, y)))\},
\] where $y_s=\{y_s^i\}_{i=1}^{N_y}$ contains totally $N_y$ logical steps in solving $x$. Previous approaches \cite{zheng2024processbench, wang2024math, lightman2023let, zhou2024your} have relied on step separation based on spacing tokens or punctuation marks. However, such methods often lead to over-segmentation due to variations in output styles among different LLMs. These styles may include specialized formatting with line breaks, spacing tokens, or symbols in mathematical equations. This issue is particularly critical if we aim to construct a universal PRM that can accommodate a wide range of policies. To address this, we prompt the model \(\pi_{\theta_s}\) with \(p_s\) to perform semantic-based step separation which can adapt to multiple output distribution. The model outputs the results in JSON format, which facilitates verification to ensure that the separation process does not alter the original content of \(y\). 

To perform the automatic labeling process in an LLM-as-a-judge manner, we leverage a high-capacity LLM $\pi_{\theta_d}$, with parameters $\theta_d$, ensuring sufficient capability for accurate judgments. We design a set of prompt strategies, $\mathcal{P}_\text{dis} = \{p_h\}_{h=1}^H$, comprising $H$ diverse prompts (Appendix \ref{app:p_dis}). Specifically, we employ two primary strategies that induce distinct inductive biases in model inference, combining them to construct multiple prompts for each query. The first strategy involves one-shot ICL with varied exemplars. By carefully selecting and altering the one-shot examples, we introduce distinct inductive biases into the LLM’s discrimination process, enabling richer perspectives on each query. The second strategy utilizes a re-reading technique \cite{xu2024re}, which adjusts the LLM's attention allocation during inference, further diversifying the inductive biases in its responses. After that, we ensemble outputs from these varied inductive biases by averaging the results. This approach can be interpreted as a form of Bayesian model averaging \cite{zhang2023and}, where the outputs from different strategies are treated as diverse yet complementary hypotheses. By ensembling outputs from these varied inductive biases, we substantially reduce over-reliance on any single strategy, mitigate prediction bias, and enhance overall performance. Additionally, we address prediction variance through a self-voting mechanism. This involves generating responses across $G$ sampling trials and selecting the majority-voted result, ensuring consistency and stability in each inference prediction. Then the overall sampling process can be performed on all training samples in the dataset $\mathcal{D}_{\text{gen}}$ generated in Section \ref{sec:universal_policy_output_generation} and can be expressed by:
\begin{equation}
\label{equ:ensemble}
\begin{aligned}
\mathcal{D}_{\text{dis}} &= \bigcup_{\mathcal{D}_{\text{gen\_sep}}} \left\{ (x, y_s, y^\star, r_s) \mid r_s = \frac{1}{H} \sum_{h=1}^{H} \text{mode}(\mathcal{R}_h^s)) \right\},
\\
\mathcal{R}_h^s &= \{r_g \sim \pi_{\theta_d}(\cdot | p_h(x, y_s, y^\star))\}_{g=1}^G,
\end{aligned}
\end{equation} where $r_s=\{r_s^i\}_{i=1}^{N_y}$ contains a list of process rewards on $N_y$ logical reasoning steps in $y_s$.

\subsubsection{Universal PRM Learning with Reverse Verification}
\label{sec:reverse_verification}

Based on the dataset \(\mathcal{D}_{\text{dis}}\) generated in Section \ref{sec:ensemble_prompting}, we proceed with training the universal PRMs. Compared to related work, our proposed framework \methodr, introduces two key differences in input information and loss function. 

Related works \cite{zheng2024processbench, lightman2023let} typically use the initial question and partial solutions as inputs, allowing the PRM to predict rewards. However, in real-world applications involving mathematical problem-solving or code generation, ground truth results are often readily available. By incorporating these ground truth results during training, our PRM can compare the final output with intermediate reasoning steps, enabling a reverse verification process. This approach can significantly enhance the PRM's ability to evaluate process rewards. Specifically, we construct the universal PRM \(R\) using a pretrained LLM \(\theta\) as the base model. The PRM predicts process rewards as follows:  
\[
\bar{r^{i}_s} = R_\theta(x, \alpha \cdot y^\star, y_s^{\leq i}), \quad y_s^{\leq i} = \{y_s^j\}_{j=1}^i,
\]  
where \(x\) is the input, \(y^\star\) is the ground truth solution, \(y_s^{\leq i}\) represents the sequence of reasoning steps before the step $i$, and $\alpha\sim\text{Bernoulli}(p_\alpha)$ controls the proportion of the ground truth outcomes provided for the reverse verification.

The second key difference lies in the choice of the loss function for optimizing the universal PRM. Unlike previous works \cite{zheng2024processbench} that rely on cross-entropy loss (as described in Equation \eqref{equ:prm_normal}), we leverage dense reward signals derived from the LLM-labeled ensemble prompting process introduced in Section \ref{sec:ensemble_prompting}. To optimize the PRM, we employ an \(L_2\)-loss function, defined per sample as:  
\[
\mathcal{L}_{\text{uPRM}} = \frac{1}{2N_y}\sum_{i=1}^{N_y}\left(r^i_{s} - \bar{r^{i}_s}\right)^2,
\]  
where \(N_y\) is the number of reasoning steps, \(r^i_s\) represents the true reward, and \(\bar{r^{i}_s}\) is the predicted reward. This loss function effectively aligns the PRM predictions with dense and soft targets, improving the robustness of the learning process. During the inference of $R_\theta$, we apply a sigmoid function to constrain the output within the range $[0,1]$. The threshold for reward prediction is determined based on validation set results, ensuring optimal predictive performance.

\begin{table*}[]
    \centering
    \caption{Following the evaluation protocol of ProcessBench \cite{zheng2024processbench}, we report the F1 scores for the respective accuracies on both erroneous and correct samples. }
    \input{table/processBench.tex}

    \label{tab:processbench}
\end{table*}

\begin{table*}[]
    \centering
    \caption{The weighted F1 scores of various models were evaluated on the UniversalBench benchmark, which encompasses seven distinct data sources. In this context, ``lng'' refers to reasoning-intensive outputs derived from long CoT policies, characterized by iterative and reflective reasoning processes aimed at thorough problem-solving. Conversely, ``shrt'' denotes the traditional shortcut-based answering approach, which emphasizes direct solutions without extensive reasoning steps.}
    \resizebox{\textwidth}{!}{
    \input{table/universalBench.tex}}
    \label{tab:universalbench}   
\end{table*}

\section{Experiments}
In this section, we evaluate the performance of the universal PRM trained using our proposed \methodr. First, we assess the effectiveness of our framework on ProcessBench \cite{zheng2024processbench}, a benchmark specifically designed to evaluate generation processes using human-annotated labels. Next, we investigate the universal capabilities of the trained PRM across diverse policy distributions. To facilitate this evaluation, we construct a novel dataset that spans a wide range of policy distributions, varying in both sequence length and step separation.

\subsection{Baselines}
To comprehensively evaluate the proposed framework, \methodr, we compare it against several open-source state-of-the-art (SOTA) models. To isolate and assess the specific improvements introduced by the \methodr framework, we also fine-tuned multiple pretrained models with different datasets that are identical to the baseline methods. This ensures that the observed differences can be attributed solely to the effectiveness of the \methodr framework.

\MS \cite{wang2024math}: Math-Shepherd is an automatic annotation framework designed to assign process labels by estimating the likelihood of each reasoning step leading to a correct final answer. The model builds upon the pretrained Mistral-7B model through fine-tuning.

\RLHF \cite{RLHFmodel}: Developed under the RLHFlow project \cite{RLHFlow}, this model leverages the annotation methodology of Math-Shepherd. It generates solutions using the Mistral-7B architecture and is fine-tuned on the Llama3.1-8B-Instruct model.

\SKY \cite{skyworkopeno12024}: Based on the Skywork open-source project \cite{wei2023skywork}, this model utilizes multi-source reasoning datasets to improve general performance in process reward prediction. It is fine-tuned on the Qwen2.5-Math-7B-Instruct model \cite{yang2024qwen25math}.

\QWEN \cite{zheng2024processbench, zhang2025lessons}: This model is directly fine-tuned on the Qwen2.5-Math-7B-Instruct model using the PRM800K dataset, which is used as a base model in ProcessBench.

\QWENPRM \cite{zhang2025lessons}: This SOTA PRM demonstrates superior performance on the ProcessBench benchmark by leveraging the LLM-as-a-judge technique to label process rewards. Unlike the proposed \methodr, this model specifically focuses on identifying and evaluating the first error occurrence in trajectories generated by Qwen-series policies.

\subsection{\UNI}

Similar to the model \QWEN and \QWENPRM, we train the \UNI model under proposed \methodr using Qwen2.5-Math-7B-Instruct as the base. Qwen2.5-Math-7B-Instruct is an instruction-tuned LLM with SOTA capabilities in mathematical reasoning. For data preparation, we leverage the NuminaMath dataset from the Numina project \cite{li2024numinamath}, sampling questions to construct the dataset $\mathcal{D}_{\text{gen}}$. To ensure diversity in policy distribution, we employ five models and two prompt templates to define the policy set $\Pi$ and prompt set $\mathcal{P}$ following Section \ref{sec:universal_policy_output_generation}. For constructing the dataset $\mathcal{D}_{\text{gen\_sep}}$, we use Qwen2.5-72B-Instruct as the reasoning steps separation model $\pi_{\theta_s}$ to perform decomposition. Subsequently, following the Section \ref{sec:ensemble_prompting}, $\mathcal{D}{\text{dis}}$ is derived from $\mathcal{D}_{\text{gen\_sep}}$ using Qwen2.5-72B-Instruct as the discriminator model $\pi_{\theta_d}$. Here, three carefully designed prompts are applied to form the discriminator-specific prompt set $\mathcal{P}_{\text{dis}}$, enabling a prompting ensemble to construct the final training dataset. To facilitate reverse verification and ensure generalization even when the ground truth outcome is unavailable, we set the sampling probability $p_\alpha=0.5$ and train the model following Section \ref{sec:reverse_verification}. This approach trains \UNI to assign process rewards for half of the samples without relying on ground truth outcomes. Full details of the training process are provided in Appendix \ref{app:training}. To ensure fairness in the evaluation process, we carefully examine and control for data contamination by verifying that none of the evaluation benchmark queries are present in the training dataset. This step prevents any overlap that could compromise the integrity of the evaluation results.

\subsection{ProcessBench Experiment}
To evaluate the generalization capabilities of PRMs, which demonstrate superior performance on benchmarks like MATH and GSM8K but often struggle with broader test cases, ProcessBench \cite{zheng2024processbench} introduces an evaluation framework based on four diverse subsets. These subsets consist of challenging mathematical problems carefully curated to encompass a wide range of question distributions. Each subset includes balanced annotations of both correct and erroneous reasoning steps, ensuring comprehensive coverage of reasoning patterns. The evaluation framework comprises 3,400 test cases, offering statistically significant insights into PRM performance. We report all the experiment results by averaging three training trials with different random seeds with the decoding parameter introduced in Appendix \ref{app:decoding}.

The experimental results presented in Table \ref{tab:processbench} show that \UNI, trained using our proposed \methodr, achieves performance on par with \QWENPRM, the SOTA PRM on the ProcessBench benchmark. These findings demonstrate the effectiveness of our approach in achieving generalization under a universal policy training distribution, while also highlighting the robustness of the ensemble prompting techniques introduced in our method.

\begin{figure*}[htbp]
\centering
\begin{minipage}[t]{0.30\textwidth}
\centering
\includegraphics[width=\textwidth]{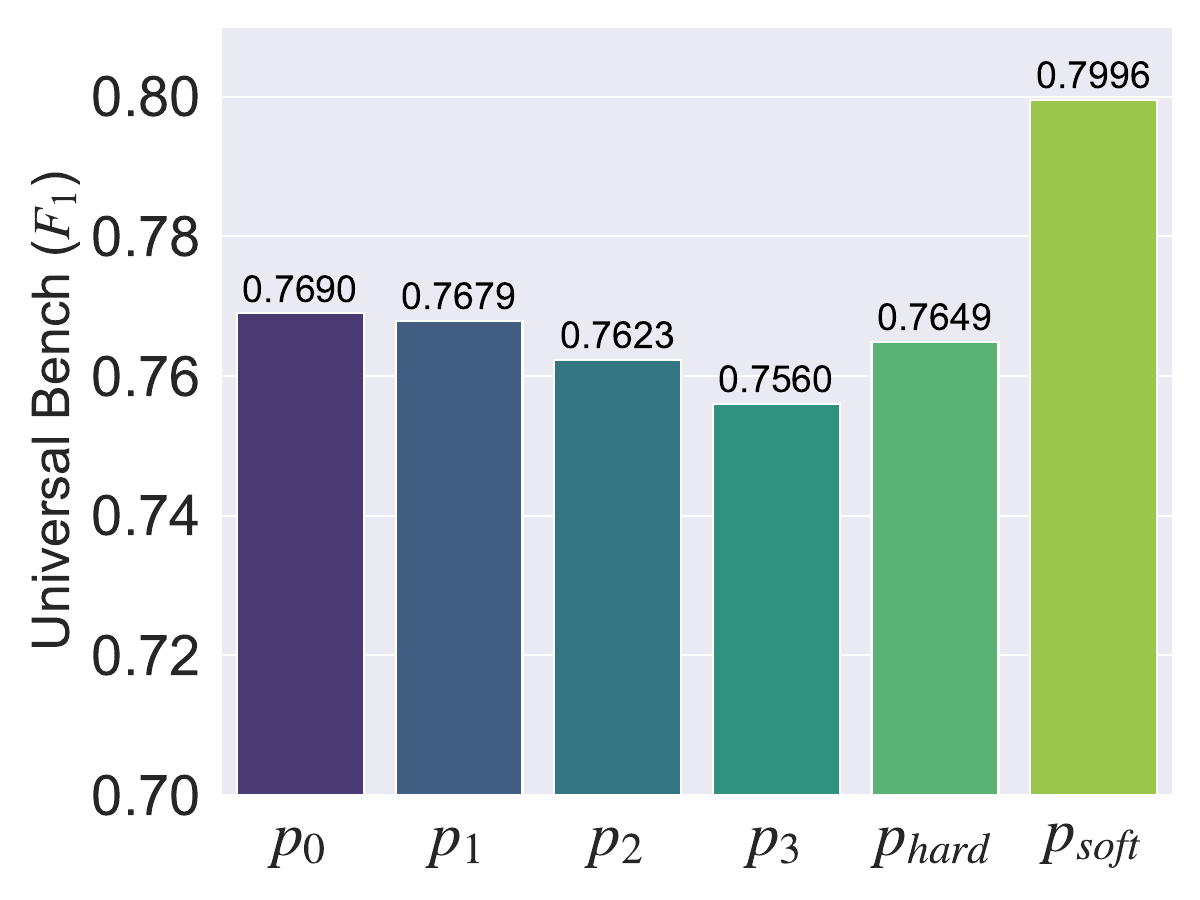}
\vspace{-0.8cm}
\caption{Ensemble Prompting}
\label{figure:enseble_prompt}
\end{minipage}
\hspace{0.025\textwidth}
\begin{minipage}[t]{0.30\textwidth}
\centering
\includegraphics[width=\textwidth]{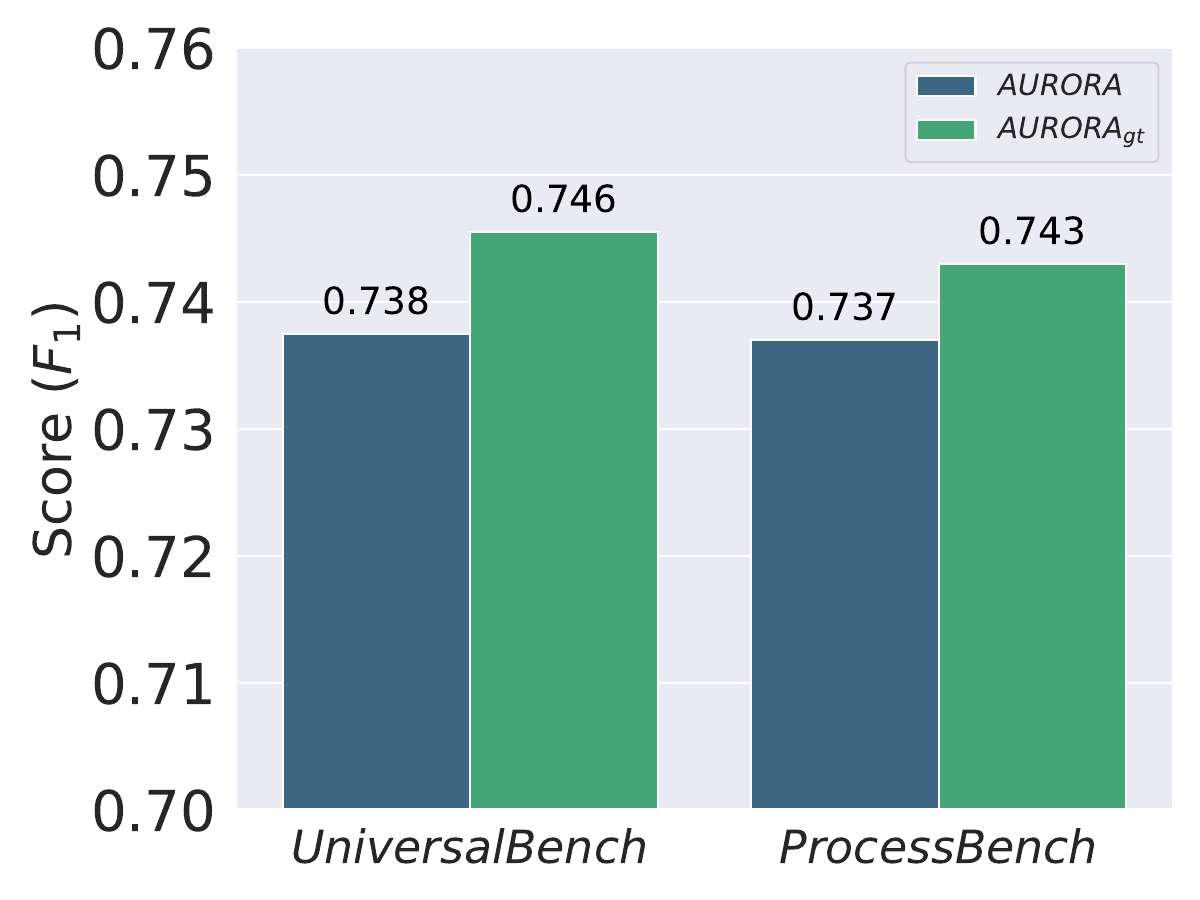}
\vspace{-0.8cm}
\caption{Reverse Verification}
\label{figure:reverse_learning_result}
\end{minipage}
\hspace{0.025\textwidth}
\begin{minipage}[t]{0.30\textwidth}
\centering
\includegraphics[width=\textwidth]{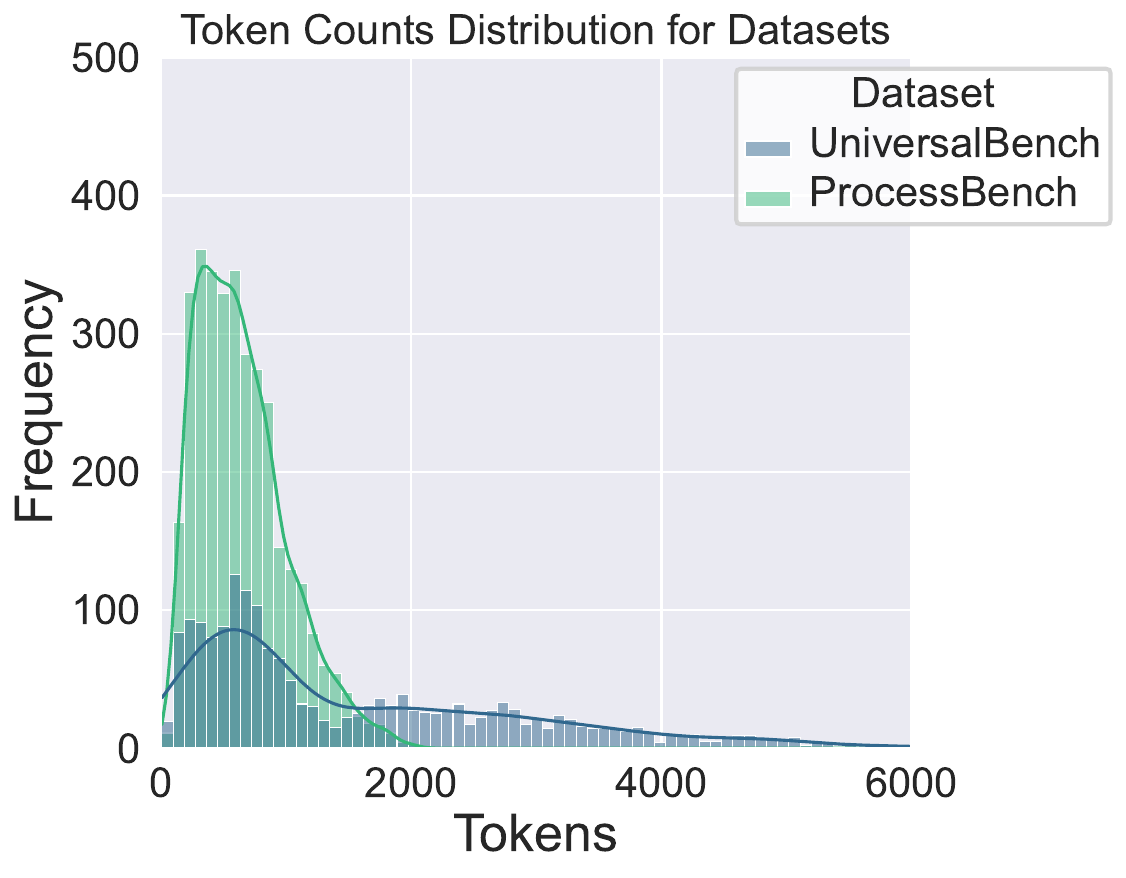}
\vspace{-0.8cm}
\caption{Token Distribution Difference}
\label{fig:token_distribution}
\end{minipage}
\vspace{-0.5cm}
\end{figure*}

\subsection{UniversalBench Experiment}

One key application of PRMs is enhancing the long CoT reasoning capabilities of LLM-based policy models. However, upon analyzing ProcessBench and evaluating performance across diverse policy distributions, particularly for outputs with long CoT styles, we observed a significant limitation: ProcessBench’s evaluation results do not fully capture the performance of the PRM on outputs with extensive token distributions. We attribute this to the fact that ProcessBench’s token distribution does not adequately cover the spectrum of long reasoning policies. 

Another important limitation of ProcessBench is its inability to fully capture performance in predicting complete reasoning trajectories, particularly those involving self-reflection and the ability to redirect reasoning from incorrect paths to correct targets. These capabilities are critical for PRMs, especially when applied in RL algorithms like Actor-Critic \cite{grondman2012survey}, where the value function approximates the expected value at each reasoning step. By focusing solely on the identification of the first error location, the evaluation method used in ProcessBench fails to effectively assess model performance in these more complex and dynamic reasoning scenarios.

To address the aforementioned challenges, we developed a new benchmark designed to encompass a broader and more representative range of policy distributions and evaluate the whole reasoning trajectory's performance. We call this benchmark UniversalBench to indicate that our proposed benchmark covers a wider range of policy distributions along the entire reasoning process, which can serve as a supplementary benchmark to ProcessBench. We also control the difficulty of the problems based on current candidate policies capability to ensure that the dataset can sufficiently explore the decision boundaries of PRMs. In this benchmark, we select GSM8K, MATH, and Olympiads as the primary benchmarks, while including IMO, AIME, and AMC as supplementary benchmarks to address the rapidly advancing mathematical reasoning capabilities of LLMs. We utilize $\Pi$, which incorporates seven distinct policies, to generate candidate responses and apply semantic reasoning step segmentation as described in Section \ref{sec:ensemble_prompting}. Human annotators are invited to carefully curate the generated responses, ensuring a balanced dataset by selecting reasoning trajectories with moderate difficulty to avoid skewed positive-negative data distributions. Following this curation process, we analyze the sequence lengths and classify the entire benchmark into seven categories based on both sequence length and source. The difference of token distribution between ProcessBench and UniversalBench is shown in Figure \ref{fig:token_distribution}. We refer the readers to Appendix \ref{app:universal_bench} for more information about UniversalBench 

Experimental results, as shown in the Table \ref{tab:universalbench}, demonstrate that \UNI training under our proposed \methodr framework has achieved superior performance, highlighting its strong generalization capabilities across diverse policy distributions. By training under the \methodr framework, \UNI effectively addresses the challenges by constructing $\mathcal{D}_{\text{gen}}$ using diverse policy and prompt distributions, particularly containing long CoT reasoning. This indicates the robustness of our approach in capturing and adapting to a wide range of policy behaviors, thereby outperforming existing methods in accuracy, making it applicable to real-world scenarios where the update policy distributions are dynamic. 

\section{Ablation Study}
\subsection{LLM-as-a-judge and Ensemble Learning}
\label{sec:ablation1}

We conduct the first ablation study to evaluate the contributions of different prompt strategies within $\mathcal{P}_{\text{dis}}$ and incorporating the ensemble technique defined in Equation \eqref{equ:ensemble}. To demonstrate the capability of automatic labeling using LLMs under our prompts with $\pi_{\theta_d}$, we follow the Section \ref{sec:ensemble_prompting} and test performance on UniversalBench. Additionally, we compare the performance of soft labels derived from the ensemble with hard labels. The results are summarized in Figure \ref{figure:enseble_prompt}. Notably, the overall performance of different strategies remains relatively stable across benchmarks but exhibits distinct performance, suggesting that different prompt strategies effectively introduce varied inductive biases when performing automatic process labeling. Ensemble hard labels bring limited improvement, which indicates that the majority-voting has effectively reduce the variance of each prompt strategies output. Ensemble labels with averaging offer a more robust and accurate estimation of the process reward. This improvement not only enhances the  accuracy of process reward estimation but also facilitates more reliable learning in the subsequent PRM training stage, leading to higher overall performance.

\subsection{Reverse Verification}
Unlike recent approaches that predict process rewards solely based on the question and solution steps, \methodr leverages ground truth outcomes to provide more accurate process rewards. This mechanism enables the PRM to verify reasoning steps in a reverse manner, leading to enhanced accuracy. To demonstrate the impact of this design, we conducted an ablation study to assess the contribution of incorporating ground truth outcomes. We also evaluated the performance of \methodr in scenarios where ground truth outcomes were unavailable. The evaluation was conducted on both ProcessBench and UniversalBench, with the results presented in Figure \ref{figure:reverse_learning_result}. The results highlight that \methodr maintains robust process reward prediction capabilities even in the absence of ground truth outcomes and achieves superior performance when ground truth outcomes are available.

\section{Conclusion}
In this paper, we introduce a novel framework, \methodr, designed for automated process reward labeling and learning using LLMs. Unlike recent approaches \cite{zheng2024processbench, lightman2023let} that operate on limited data distributions, rely solely on questions and partial solutions, or focus only on the first error occurrence, \methodr aims to train universal PRMs by addressing these limitations. Specifically, \methodr collects candidate reasoning trajectories from diverse policy distributions, evaluates process rewards across the entire reasoning sequence to support downstream RL algorithms, and incorporates reverse verification and ensemble prompting techniques to further enhance performance. To comprehensively evaluate our approach, we curated a new benchmark, UniversalBench, which captures a wide range of policy distribution and especially contains long CoT policy outputs that closely mirror real-world PRM usage scenarios in optimizing long CoT policies. Experiments on both ProcessBench and UniversalBench demonstrate that \UNI, trained using \methodr, achieves SOTA performance. We have open-sourced \UNI\footnotemark[\value{footnote}] and will release UniversalBench to encourage community adoption and further research.

\section{Impact Statements}
In this paper, we present a novel framework, called \methodr, designed to automatically construct PRMs using ensemble prompting and reverse verification. To support easy replication and foster community engagement, all candidate policies and base datasets used in this project are open-sourced, without relying on LLM API services that may undergo frequent version changes. Additionally, the trained PRM, \UNI, is also open-sourced, allowing for straightforward replication using ProcessBench and using it for subsequent RL training and guided search algorithms. As part of this effort, we introduce a new benchmark, UniversalBench, to evaluate performance across a broader range of policy distributions, which will be made available as part of the project. We believe this work is both rigorous and valuable to the field of PRMs. This paper is entirely original and has not been distributed or reviewed elsewhere. This work will be open-sourced at \url{auroraprm.github.io}, which is also following the anonymous policy during the reviewing phase. 

\bibliography{ref}
\bibliographystyle{ref_sty}


\newpage
\appendix
\onecolumn

\section{Training Details of \UNI}
\label{app:training}

\UNI is fine-tuned based on Qwen2.5-Math-7B \cite{yang2024qwen25math}. The original language modeling head is replaced with a reward head designed to output a one-dimensional reward score. The model strictly follows the chat template of the base model and does not incorporate any additional special tokens. The loss function is calculated exclusively at the position of the final token in each training step. The queries used for generating the training dataset were sourced from Numina, 690,000 samples is generated using the \methodr framework. The fine-tuning process employs the loss function of the mean squared error (MSE), with a learning rate of 1e-6 and a batch size of 16, over 1.5 epochs.

\section{Decoding Parameters}
\label{app:decoding}

In our step segmentation process, we implemented a greedy decoding strategy to enforce deterministic and consistent step boundaries. For all other stages, including answer generation and LLM-based verification, we employed a stochastic decoding approach utilizing nucleus sampling (Top-p = 0.85) with a temperature setting of 0.7. This configuration facilitated controlled diversity while preserving response coherence. To enhance robustness and mitigate variance, multiple samples were generated at each stage. This approach strikes a balance between exploration and stability, enabling the model to capture a broader spectrum of reasoning paths while ensuring the reliability of the generated outputs.

\section{UniversalBench}
\label{app:universal_bench}

We curated mathematical reasoning problems from seven distinct sources: GSM8K \cite{cobbe2021gsm8k}, MATH \cite{hendrycksmath2021}, Olympiad benchmark \cite{he2024olympiadbenchchallengingbenchmarkpromoting}, AIME, AMC, and IMO. This comprehensive collection spans 662 problems ranging from elementary school level to advanced competition difficulty, ensuring broad coverage of mathematical reasoning challenges.\\
To generate answers, we utilized seven open-source models with sizes ranging from 7 billion to 72 billion parameters. These include Qwen2.5-72B-Instruct \cite{qwen2.5}, Yi-34B-Instruct \cite{ai2024yi}, INF-34B-Instruct \cite{inf-llm}, Llama3.1-70B-Instruct, Llama3.1-8B-Instruct \cite{dubey2024llama}, as well as two additional long chain-of-thought (COT) policies, Qwen2.5-32B-QwQ \cite{teamqwen2024qwq} and INF-o1-$\pi_0$ \cite{inftech_pi_zero2024}, specifically chosen for their superior performance on challenging competition questions. This inclusion enhances the benchmark's universality. For benchmarks targeting more challenging problems such as AIME, AMC, and IMO, we filtered out answers generated by policies with lower accuracy. Conversely, for less challenging benchmarks like MATH and GSM8K, answers produced by long COT models were also filtered. The Olympiad benchmark is unique due to its high difficulty but large dataset size, and since most models have been extensively trained on this dataset, both long COT and shortcut policies' outputs were retained. Given that long COT policies produce more solution steps and offer a broader search space, five answers were sampled per problem to enhance coverage. After filtering out cases with abnormal reasoning, a total of 1849 question-answer pairs were generated.\\
Given the diversity of policy models, it was challenging to establish a unified step segmentation marker. Considering the inefficiency and high cost associated with manual annotation, we adopted an LLM-based reformatting method aimed at dividing answers into sequences of roughly equal granularity. We designed prompts instructing the LLM to segment the answers logically without altering their content. Utilizing Qwen2.5-72B-Instruct, we performed segmentation while ensuring consistency in answer length before and after the process, followed by human verification for accuracy. However, for long COT policies where steps often number in the hundreds, maintaining consistency between the number of steps and their content proved difficult. Therefore, for these outputs, we used double line breaks as a delimiter and merged every five steps into one to balance segmentation rationality and annotation feasibility. Post-processing resulted in an average of 13.7 steps per answer in the universal benchmark, compared to 9.3 for shortcut answers and 19.1 for long COT answers.\\
A group of human experts with competitive math experience was hired to annotate the dataset. To ensure their proficiency, we assessed their mathematical abilities through simple problems before allowing them to annotate. Prior to official annotation tasks, two experts independently annotated the same subset to ensure a satisfactory level of agreement, which was also applied during the final quality check phase. This rigorous approach ensured the reliability and accuracy of the annotations.

\section{Prompt Set $\mathcal{P}_{\text{ gen}}$ }
\label{app:p_gen}
\begin{table}[H]
    \centering
    \caption{Policy Models Overview}
    \begin{tabular}{lc}
        \toprule
        \textbf{Policy Model} & \textbf{Size (Billion Parameters)} \\
        \midrule
        Qwen2.5-72B-Instruct   & 72B \\
        Yi-34B-Instruct        & 34B \\
        Llama3.1-70B-Instruct  & 70B \\
        INF-34B-Instruct    & 34B \\
        Llama3.1-8B-Instruct   & 8B  \\
        Qwen2.5-32B-QwQ & 32B \\
        INF-o1-$\pi_0$ & 32B \\
        \bottomrule
    \end{tabular}
\end{table}

The prompt as illustrated by the "QwQ system prompt" is utilized for answer generation in the Qwen2.5-32B-QwQ model, whereas the "INF-o1 {$\pi_0$} system prompt" is employed for generating responses in the INF-o1 $\pi_0$ model. All other policy models utilize the "default system prompt" for their respective response generations.

\begin{tcolorbox}[title=Default system prompt, label={fig:generate_prompt_default}, breakable]
\textbf{[System]:}\\
You are a helpful assistant.
\end{tcolorbox}

\begin{tcolorbox}[title=QwQ system prompt, label={fig:generate_prompt_qwq}, breakable]
\textbf{[System]:}\\
You are a helpful and harmless assistant. You are Qwen developed by Alibaba. You should think step-by-step.
\end{tcolorbox}

\begin{tcolorbox}[title=INF-o1 {$\pi_0$} \text{ system prompt}, label={fig:generate_prompt_infpi0}, breakable]
\textbf{[System]:}\\
You are an advanced AI language model specializing in solving math and programming problems step by step. Carefully analyze each part of the problem, verify the accuracy of your reasoning with relevant facts and data, and provide clear, logical solutions. Reflect on and review your approach throughout the problem-solving process to ensure precision and thoroughness. Always think through the problem step by step and provide your answers accordingly.
\end{tcolorbox}

During the answer generation process, a randomly selected question prompt is assigned to each question to enhance the diversity of the answer set.
\begin{tcolorbox}[title=Question prompt $p_{0}$, label={fig:generate_question_prompt}, breakable]
\textbf{[User]:}\\
\textcolor{red}{\{question\}}
\end{tcolorbox}
\begin{tcolorbox}[title=Question prompt $p_{1}$, label={fig:generate_question_prompt}, breakable]
\textbf{[User]:}\\
\textcolor{red}{\{question\}}\\
Let's think step by step.\\
\end{tcolorbox}
\begin{tcolorbox}[title=Question prompt $p_{2}$, label={fig:generate_question_prompt}, breakable]
\textbf{[User]:}\\
\textcolor{red}{\{question\}}\\
First, deeply analyze the problem and identify key concepts and relationships, then solve it step by step with clear reasoning.\\
\end{tcolorbox}

\section{Reasoning Step Separation Prompt $p_s$}
\label{app:p_s}
\input{appendix_prompt/seperate_prompt}
\section{Prompt Set $\mathcal{P}_{\text{dis}}$}
\label{app:p_dis}
\input{appendix_prompt/verify_v01}
\input{appendix_prompt/verify_v01_1_shot}
\input{appendix_prompt/verify_v01_rr}
\input{appendix_prompt/verify_v1_1shot}


\end{document}

%% file: algo.tex
\newcommand{\method}{\textsc{AURORA}\xspace}

\newcommand{\algofont}[1]{{\scalefont{1.0}{\texttt{\textbf{#1}}}}}

\newcommand{\MScolor}{blue!60!green}
\newcommand{\MSnc}{\algofont{Math-Shepherd-PRM-7B}}
\newcommand{\MS}{{\color{\MScolor}\MSnc}\xspace}

\newcommand{\RLHFcolor}{green!60!black}
\newcommand{\RLHFnc}{\algofont{RLHFlow-PRM-Mistral-8B}}
\newcommand{\RLHF}{{\color{\RLHFcolor}\RLHFnc}\xspace}

\newcommand{\SKYcolor}{gray}
\newcommand{\SKYnc}{\algofont{Skywork-PRM-7B}}
\newcommand{\SKY}{{\color{\SKYcolor}\SKYnc}\xspace}

\newcommand{\QWENcolor}{magenta!80!black}
\newcommand{\QWENnc}{\algofont{Qwen2.5-Math-7B-PRM800K}}
\newcommand{\QWEN}{{\color{\QWENcolor}\QWENnc}\xspace}

\newcommand{\QWENPRMcolor}{red!60!black}
\newcommand{\QWENPRMnc}{\algofont{Qwen2.5-Math-PRM-7B}}
\newcommand{\QWENPRM}{{\color{\QWENPRMcolor}\QWENPRMnc}\xspace}

\newcommand{\UNIcolor}{red!60!orange}
\newcommand{\UNInc}{\algofont{Universal-PRM-7B}}
\newcommand{\UNI}{{\color{\UNIcolor}\UNInc}\xspace}

\newcommand{\methodnc}{\algofont{\method}}
\newcommand{\methodr}{{\color{\UNIcolor}\methodnc}\xspace}

%% file: table/processBench.tex
\begin{tabular}{lccccc}
\bottomrule
\textbf{Model} & 
\textbf{GSM8K} & 
\textbf{MATH} & 
\makecell{\textbf{Olympiad-} \\ \textbf{Bench}} & 
\makecell{\textbf{Omni-} \\ \textbf{MATH}} &
\textbf{Average} \\
\midrule
\MS & 47.9 & 29.5 & 24.8 & 23.8 & 31.5 \\
\RLHF & 50.4 & 33.4 & 13.8 & 15.8 & 28.4 \\
\SKY & 70.8 & 53.6 & 22.9 & 21.0 & 42.1 \\
\QWEN & 68.2 & 62.6 & 50.7 & 44.3 & 56.5 \\
\QWENPRM & 82.4 & 77.6 & 67.5 & 66.3 & 73.5 \\
\midrule
\UNI & \textbf{85.8} & \textbf{77.7} & \textbf{67.6} & \textbf{66.4} & \textbf{74.3} \\
\bottomrule
\end{tabular}

%% file: table/universalBench.tex
\begin{tabular}{lcccccccc}
\bottomrule
\textbf{Model} & 
\makecell{\textbf{AIME} \\ \textbf{(lng)}} & 
\makecell{\textbf{AMC} \\ \textbf{(lng)}} & 
\makecell{\textbf{IMO} \\ \textbf{(lng)}} & 
\makecell{\textbf{Olympiads} \\ \textbf{(lng)}} & 
\makecell{\textbf{GSK8K} \\ \textbf{(shrt)}} & 
\makecell{\textbf{Olympiads} \\ \textbf{(shrt)}} & 
\makecell{\textbf{MATH} \\ \textbf{(shrt)}} & 
\textbf{Average}\\
\midrule
\MS & \textbf{60.0} & 14.1 & 57.6 & 49.3 & 40.8 & 24.3 & 43.9 & 41.4\\
\RLHF & 18.7 & 34.6 & 23.7 & 11.3 & 72.1 & 45.0 & 56.8 & 37.4\\
\SKY & 24.0 & 13.2 & 21.8 & 16.5 & 33.9 & 61.7 & 31.8 & 28.9\\
\QWEN & 57.1 & 56.8 & \textbf{65.4} & 54.9 & 89.6 & 74.0 & 81.9 & 68.5\\
\QWENPRM & 49.0 & 61.6 & 45.3 & 60.2 & 88.8 & 73.7 & 80.7 & 65.6\\
\midrule
\UNI & 59.5 & \textbf{76.2} & 62.8 & \textbf{65.5} & \textbf{91.9} & \textbf{80.2} & \textbf{85.8} & \textbf{74.5}\\
\bottomrule
\end{tabular}

%% file: appendix_prompt/seperate_prompt.tex
\begin{tcolorbox}[title=LLM discrimination prompt $p_s$, label={fig:verify_prompt_v01_rr}, breakable]
\textbf{[System]:}\\
You are a helpful assistant who can seperate the logical steps accurately.
\tcblower
\textbf{[User]:}\\

Please split the following math problem-solving text according to logical 
steps, generating a JSON object where each step is an independent key-value 
pair in the format `\{\{ "Step X": "Content of the step" \}\}`, where `X` is the step number. Note:\\

\# Rules:\\
1. Retain the original text for each step without any modifications, additions, or deletions, only splitting based on logical steps.\\
2. If the text does not contain explicit step indications, split the steps according to the logical flow of the content. Each step should have a conclusion progression relative to the previous step and represent a complete intermediate conclusion (e.g., equations, intermediate results, planning thoughts, etc.).\\
3. The final output should be a JSON object containing each logical step.\\
\\
\textcolor{red}{\{shot\}}
\\
**Text to process:**\\
\\
\textcolor{red}{\{answer\}}
\\
Please output directly according to the above instructions.\\

\end{tcolorbox}

%% file: appendix_prompt/verify_v01.tex
\begin{tcolorbox}[title= LLM discrimination prompt $p_0$, label={fig:verify_prompt_v01_sp}, breakable]

\textbf{[System]:}\\
Your Role and Task:\\
\\
You are a math teacher, and I need your help in grading exams. I will provide you with the question, standard answer, and student answer. Based on the standard answer, you need to determine the correctness of each step in the student’s answer. The student answer will be given in JSON format, detailing each step of their solution, and you should indicate whether each step is correct or incorrect.\\
\\
Important Notes:\\
1. The student’s answer may have a different approach from the standard answer. If the student’s reasoning is logically sound and their final answer matches the standard answer, then it should be considered correct.\\
2. You need to assess each step’s correctness and mark it with 0 for incorrect and 1 for correct. For example, if there are three steps, where the first is correct, and the second and third are incorrect, then at the end of your response, output a list named judge\_result like this: judge\_result=[1,0,0].The judge\_result can only contain 0 and 1.\\
3. If a step (step i) is incorrect because of an error in the previous step (step i-1), it should be considered wrong as well, even if the deduction or calculation in step i itself is technically correct.\\
4. If a step references an unrelated or inapplicable conclusion (even if the conclusion itself is correct), that step should be considered incorrect.\\
5. Critically evaluate all steps and results in your answer, making sure each step has an evaluation conclusion.\\
\\
The user will provide the question, standard answer, and student answer. Please grade the student answer strictly according to these instructions and include the final judge\_result list at the end of your response, like this format: judge\_result=[1,0,0]
\tcblower
\textbf{[User]:}\\
\# question:\\
\textcolor{red}{\{question\}}\\
\\
\# standard answer:\\
\textcolor{red}{\{ground\_truth\_solution\}}\\
\\
\# student answer:\\
\textcolor{red}{\{student\_solution\}}\\
\\
\# your output:\\

\end{tcolorbox}

%% file: appendix_prompt/verify_v01_1_shot.tex
\begin{tcolorbox}[title=LLM discrimination prompt $p_1$, label={fig:verify_prompt_v01_rr}, breakable]
\textbf{[System]:}\\
\# Your Role and Task:\\
\\
You are a math teacher, and I need your help in grading exams. I will provide you with the question, standard answer, and student answer. Based on the standard answer, you need to determine the correctness of each step in the student’s answer. The student answer will be given in JSON format, detailing each step of their solution, and you should indicate whether each step is correct or incorrect.\\
\\
\# Important Notes:\\
1. The student’s answer may have a different approach from the standard answer. If the student’s reasoning is logically sound and their final answer matches the standard answer, then it should be considered correct.\\
2. You need to assess each step’s correctness and mark it with 0 for incorrect and 1 for correct. For example, if there are three steps, where the first is correct, and the second and third are incorrect, then at the end of your response, output a list named judge\_result like this: judge\_result=[1,0,0].\\
3. If a step (step i) is incorrect because of an error in the previous step (step i-1), it should be considered wrong as well, even if the deduction or calculation in step i itself is technically correct.\\
4. If a step references an unrelated or inapplicable conclusion (even if the conclusion itself is correct), that step should be considered incorrect.\\
5. Critically evaluate all steps and results in your answer, making sure each step or results has an evaluation conclusion.\\
\\
The user will provide questions, model answers, and student answers. Please follow these instructions carefully to grade the student answers. You only need to respond to the final Judge\_result list, as example: judge\_result=[1,1,1,1].Do not do any additional explanation.
\\
\textcolor{red}{\{shot\}}\\
\tcblower

\textbf{[User]:}\\
\# question:\\
\textcolor{red}{\{question\}}\\
\\
\# standard answer:\\
\textcolor{red}{\{ground\_truth\_solution\}}\\
\\
\# student answer:\\
\textcolor{red}{\{student\_solution\}}\\
\\
\# your output:\\

\end{tcolorbox}

%% file: appendix_prompt/verify_v01_rr.tex
\begin{tcolorbox}[title=LLM discrimination prompt $p_2$, label={fig:verify_prompt_sp}, breakable]
\textbf{[System]:}\\
Your Role and Task:\\
\\
You are a math teacher, and I need your help in grading exams. I will provide you with the question, standard answer, and student answer. Based on the standard answer, you need to determine the correctness of each step in the student’s answer. The student answer will be given in JSON format, detailing each step of their solution, and you should indicate whether each step is correct or incorrect.\\
\\
Important Notes:\\
1. The student’s answer may have a different approach from the standard answer. If the student’s reasoning is logically sound and their final answer matches the standard answer, then it should be considered correct.\\
2. You need to assess each step’s correctness and mark it with 0 for incorrect and 1 for correct. For example, if there are three steps, where the first is correct, and the second and third are incorrect, then at the end of your response, output a list named judge\_result like this: judge\_result=[1,0,0].The judge\_result can only contain 0 and 1.\\
3. If a step (step i) is incorrect because of an error in the previous step (step i-1), it should be considered wrong as well, even if the deduction or calculation in step i itself is technically correct.\\
4. If a step references an unrelated or inapplicable conclusion (even if the conclusion itself is correct), that step should be considered incorrect.\\
5. Critically evaluate all steps and results in your answer, making sure each step has an evaluation conclusion.\\
\\
The user will provide the question, standard answer, and student answer. Please grade the student answer strictly according to these instructions and include the final judge\_result list at the end of your response, like this format: judge\_result=[1,0,0]
\tcblower
\textbf{[User]:}\\
\# question:\\
\textcolor{red}{\{question\}}\\
\\
\# standard answer:\\
\textcolor{red}{\{ground\_truth\_solution\}}\\
\\
\# student answer:\\
\textcolor{red}{\{student\_solution\}}\\
\\
Your Role and Task:\\
\\
You are a math teacher, and I need your help in grading exams. I will provide you with the question, standard answer, and student answer. Based on the standard answer, you need to determine the correctness of each step in the student’s answer. The student answer will be given in JSON format, detailing each step of their solution, and you should indicate whether each step is correct or incorrect.\\
\\
Important Notes:\\
1. The student’s answer may have a different approach from the standard answer. If the student’s reasoning is logically sound and their final answer matches the standard answer, then it should be considered correct.\\
2. You need to assess each step’s correctness and mark it with 0 for incorrect and 1 for correct. For example, if there are three steps, where the first is correct, and the second and third are incorrect, then at the end of your response, output a list named judge\_result like this: judge\_result=[1,0,0].The judge\_result can only contain 0 and 1.\\
3. If a step (step i) is incorrect because of an error in the previous step (step i-1), it should be considered wrong as well, even if the deduction or calculation in step i itself is technically correct.\\
4. If a step references an unrelated or inapplicable conclusion (even if the conclusion itself is correct), that step should be considered incorrect.\\
5. Critically evaluate all steps and results in your answer, making sure each step has an evaluation conclusion.\\
\\
The user will provide the question, standard answer, and student answer. Please grade the student answer strictly according to these instructions and include the final judge\_result list at the end of your response, like this format: judge\_result=[1,0,0]

\end{tcolorbox}

%% file: appendix_prompt/verify_v1_1shot.tex
\begin{tcolorbox}[title=LLM discrimination prompt $p_3$, label={fig:verify_prompt_sp}, breakable]

\textbf{[System]:}\\
\# Your Role and Task:\\
\\
You are a math teacher, and I need your help in grading exams. I will provide you with the question, standard answer, and student answer. Based on the standard answer, you need to determine the correctness of each step in the student’s answer. The student answer will be given in JSON format, detailing each step of their solution, and you should indicate whether each step is correct or incorrect.\\
\\
\# Important Notes:\\
1. The student’s answer may have a different approach from the standard answer. If the student’s reasoning is logically sound and their final answer matches the standard answer, then it should be considered correct.\\
2. You need to assess each step’s correctness and mark it with 0 for incorrect and 1 for correct. For example, if there are three steps, where the first is correct, and the second and third are incorrect, then at the end of your response, output a list named judge\_result like this: judge\_result=[1,0,0].\\
3. If a step (step i) is incorrect because of an error in the previous step (step i-1), it should be considered wrong as well, even if the deduction or calculation in step i itself is technically correct.\\
4. If a step references an unrelated or inapplicable conclusion (even if the conclusion itself is correct), that step should be considered incorrect.\\
5. Critically evaluate all steps and results in your answer, making sure each step or results has an evaluation conclusion.\\
\\
\textcolor{red}{\{shot\}}\\
\\
The user will provide questions, model answers, and student answers. Please follow these instructions carefully to grade the student answers. You only need to respond to the final judge\_result list, as example: judge\_result=[1,1,1,1].Do not do any additional explanation.
\tcblower
\textbf{[User]:}\\
\# question:\\
\textcolor{red}{\{question\}}\\
\\
\# standard answer:\\
\textcolor{red}{\{ground\_truth\_solution\}}\\
\\
\# student answer:\\
\textcolor{red}{\{student\_solution\}}
\end{tcolorbox}